\documentclass[runningheads]{llncs}
\usepackage{graphicx}
\usepackage{makecell}
\usepackage{hyperref}
\usepackage{amsmath,amssymb,amsfonts} 

\usepackage{ifthen}
\usepackage{xcolor}

\begin{document}
\title{Enhancing exploration algorithms for navigation with visual SLAM\thanks{This research was supported by the Ministry of Science and Higher Education of the Russian Federation, project No 075-15-2020-799}}
%
%
\author{Kirill Muravyev\inst{1,2}\orcidID{0000-0001-5897-0702} \and Andrey Bokovoy\inst{1}\orcidID{0000-0002-5788-5765} \and Konstantin Yakovlev\inst{1,2}\orcidID{0000-0002-4377-321X}}
%
%
\institute{Artificial Intelligence Research Institute, Federal Research Center for Computer Science and Control of Russian Academy of Sciences, Moscow, Russia.
\and
Moscow Institute of Physics and Technology, Dolgoprudny, Russia
\email{\{bokovoy,yakovlev\}@isa.ru, kirill.mouraviev@yandex.ru}}
%
\maketitle              
\begin{abstract}
Exploration is an important step in autonomous navigation of robotic systems. In this paper we introduce a series of enhancements for exploration algorithms in order to use them with vision-based simultaneous localization and mapping (vSLAM) methods. We evaluate developed approaches in photo-realistic simulator in two modes: with ground-truth depths and neural network reconstructed depth maps as vSLAM input. We evaluate standard metrics in order to estimate exploration coverage.

\end{abstract}
\section{Introduction}

Making robotic systems fully autonomous is an important problem for modern researchers~\cite{gonzalez2017supervisory,papachristos2019autonomous,tang2019topological,choi2020development}. In order to operate autonomously, the system needs to know it's position on the map (environment). If the environment is unknown, the map is need to be built first. This step can be done by manually controlling the robotic system, however that's not always possible due to operating conditions, e.g. poor signal for remote control. So, one way of solving this problem is using Simultaneous Localization and Mapping (SLAM) with exploration algorithms.

SLAM algorithms are used to build the map of an unknown environment and retrieve robot's current position by utilizing different sensors. There are a lot of different sensors for SLAM to operate with, such as: GPS~\cite{burschka2004v}, lidar~\cite{droeschel2018efficient,hening20173d}, inertia measurement unit (IMU)~\cite{nutzi2011fusion,vidal2018ultimate} and cameras~\cite{lemaire2007vision,asadi2018vision}. The choice of the sensors is done considering the operating environment, robot's size or weight restrictions and etc. 

In GPS-denied environment, popular sensor of choice is monocular camera. Modern monocular vSLAM algorithms use convolutional neural networks to reconstruct depth maps~\cite{tateno2017cnn,bokovoy2019real}. Those depth maps are suitable as an input for RGB-D vSLAM algorithms. However, these algorithms still suffer from problems that are common for all vSLAM algorithms: incorrect scale, localization errors during rotations without translation and poorly detailed environment.

The second important part of autonomous navigation is \textbf{exploration}~\cite{burgard2000collaborative,sim2005global}. At each step of localization and mapping the algorithms decide where robot needs to go in order to explore more unknown space and map it. Modern algorithms are versatile and usually work with 2D maps and poses (SLAM output). However, to increase the robustness of autonomous navigation with vision-based SLAM, we need to consider vSLAM problems in exploration algorithms. In this work we introduce such enhancements and evaluate them in photo-realistic simulated environment.

This paper is organized as follows: section \ref{sec:related_work} describes current state of research in visual SLAM and autonomous exploration. Section \ref{sec:problem_statement} states the exploration problem formally. Section \ref{sec:method_overview} describes proposed exploration pipeline detailly. Section \ref{sec:experiments} presents the experimental setup and the results of the experiments in both RGB-D and monocular modes. Section \ref{sec:conclusion} concludes.

\section{Related work}
\label{sec:related_work}

Exploration is crucial for autonomous navigation in unknown environment, so there exists a vast variety of methods and algorithms aimed at solving this task. We focus on methods that work in conjunction with visual SLAM or work with visual sensors. This section gives brief overview.

In early works authors used known information about operating environment. For example, in~\cite{santosh2008autonomous} knowledge about geometric forms of floor, walls, ceiling and etc. is used in order to extract frontiers directly from the images. The goal point is chosen at the most informative place on the map based on the extracted information. In order to find the shortest path, Dijkstra's pathplanning algorithms is used. The algorithm is tested on real robot in indoor environment, so the developed visual servo control is applied in order to reach the destination. Regardless being able to solve small and medium-scale exploration tasks, this approach is applicable to indoor exploration only and not robust to environment changes and large scale exploration.

In~\cite{dayoub2013vision} authors use exploration with graph-based stereo SLAM. However, in order to perform shortest pathplanning, local semi-continuous metric space is used in opposite to following the nodes of the graph. Authors also implemented visual odometry failure recovery in order to improve the final quality of the localization and exploration. The algorithm is able to navigate autonomously for 30 minutes (limited by robot's battery) in real indoor environment.

Another approach that utilizes monocular vSLAM is presented in~\cite{von2016autonomous}. Semi-dense LSD-SLAM~\cite{engel2014lsd} algorithm is used for mapping and localization of micro aerial vehicle (MAV). As an exploration algorithm authors introduce star discovery. This approach is used in order to overcome the visual odometry drift and errors during rotation without motion (common problem for monocular vSLAM). The exploration algorithm is pretty straightforward: MAV performs star discovery on spot, then the farthest point on the frontier in line of sight is chosen as a new star discovery point, then the robot proceeds to this point. This approach require a lot of free space in order to perform a star rotation. That limits the application area of this algorithm drastically.  

More recent approach~\cite{gao2018improved} introduces an enhancement to frontier search by making use of heading information and coarse graph representation of the map in order to improve map coverage and reduce time of exploration. The algorithm is tested on wheeled robot platform. Authors report that the robot were able to explore large office ($250m^2$) with different obstacles in 7 minutes. Regardless this algorithm originally works with laser scanner only, the introduced approach can be adopted for vSLAM autonomous exploration.

\section{Problem statement}
\label{sec:problem_statement}

The exploration problem that we consider is described as follows. A robot equipped with only visual sensors (monocular, stereo or RGB-D camera) is located in unknown environment of restricted area (usually indoor space). Its task is to construct a 2D map of whole environment while moving through it.

At each step $t$, a robot is given by observation $I_t$ - an image from its camera. Using this observation, exploration algorithm tracks its location, maps information from the observation and decides where to move to explore and map new space. The output of the algorithm $A$ at step $t$ is $M_t$ - a map of explored part of environment, and an action $a_t$ - an intention to move somewhere:

$$A(I_t, M_{t-1}) = (M_t, a_t)$$

A map is represented as 2D matrix and consists of free, occupied and unexplored cells. Each cell of this matrix represents a small square of fixed size (e.g. 5x5 cm). The matrix is also provided with position of its top-left corner in global coordinate system. At initial step, the map is an empty matrix.

$$M_t = (P_t \in \{0, 1, -1\}^{H \times W}; (x_t, y_t) \in \mathbb{R}^2);\ \ M_0 = (\emptyset; (0, 0))$$

An action is represented as a robot pose shift:

$$a_t = (dx, dy, \delta)$$

It is the command for robot "move on distance $(dx, dy)$ (relatively to its current position) and rotate by angle $\delta$". To simplify our model, we consider only four possible actions: move forward ($a_t = (dx, 0, 0)$), turn left ($a_t = (0, 0, \delta)$), turn right ($a_t = (0, 0, -\delta)$), and remain on the spot ($a_t = (0, 0, 0)$).

To measure exploration efficiency, absolute and relative coverage metric are commonly used. The value of the absolute metric is the area of explored map at certain time steps $t$. The value of the relative metric is the percentage of environment space that has been explored at certain time steps $t$:

\begin{equation}
    C_{abs} = \{|(i,j): M_t^{i,j} \geq 0|\}, t \in T
    \label{eq_cov_abs}
\end{equation}

\begin{equation}
    C_{rel} = \lbrace \frac{|(i,j): M_t^{i,j} \geq 0|}{|(i,j): M^{i,j} \geq 0|} \rbrace, t \in T
    \label{eq_cov_rel}
\end{equation}

where $M$ is ground-truth map of the whole environment.

\section{Method overview}
\label{sec:method_overview}

\begin{figure}
    \centering
    \includegraphics[width=0.8\textwidth]{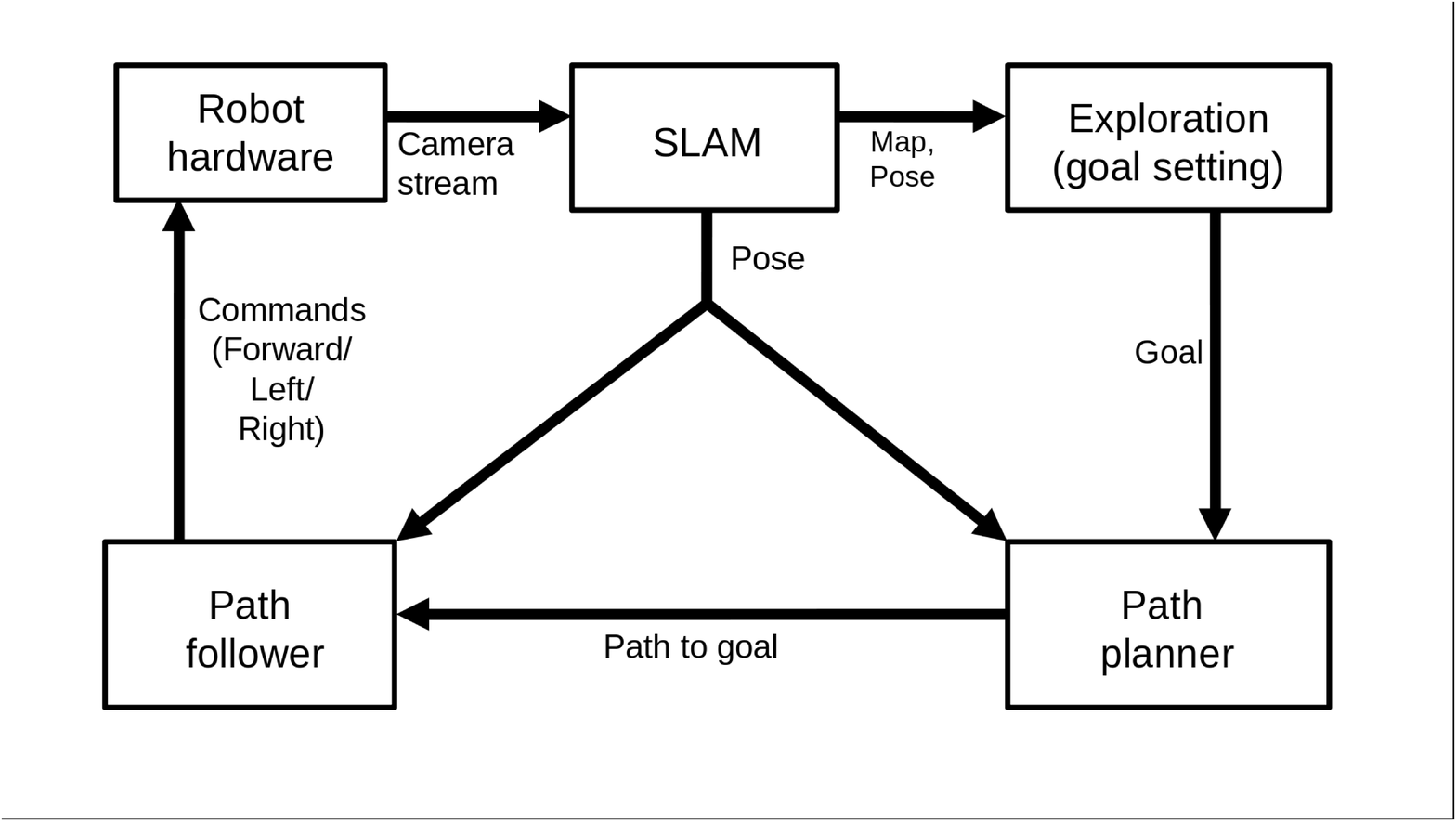}
    \caption{A scheme of proposed exploration pipeline}
    \label{figure_pipeline}
\end{figure}

We propose fully autonomous exploration pipeline for robots equipped with visual sensors. Our pipeline consists of four parts:

\begin{itemize}
    \item \textbf{SLAM} module takes data from robot's camera and estimates its trajectory and 2D map of environment simultaneously in real time
    \item \textbf{Exploration} module takes current estimated robot position and SLAM-builded map, and chooses goal where the robot should go
    \item \textbf{Path planning} module builds path from robot to goal position in SLAM-produced map
    \item \textbf{Path following} module takes current robot position and path to goal and sets low-level commands to robot's controller: where to move now - forward, left, or right
\end{itemize}

To simplify implementation on real robots and interaction between modules, we integrate our pipeline with Robot Operation System (ROS)\footnote{http://www.ros.org}. Full scheme of the pipeline is shown at figure \ref{figure_pipeline}.

\subsection{SLAM}

We chose RTAB-MAP algorithm \cite{labbe2019rtab} to perform real-time simultaneous localization and mapping. Our choice is motivated by RTAB-MAP has open-source ROS implementation\footnote{http://wiki.ros.org/rtabmap\_ros} and has wide range of adjustable parameters. It takes stereo or RGB-D images as input and outputs robot's trajectory and map of environment in both 2D and 3D. Trajectory is stored as a set of 6 DoF poses, 2D map is stored as occupancy grid (a matrix of free, occupied and unknown cells), and 3D map is stored as point cloud.

Off-the-shelf RTAB-MAP algorithm works only with stereo or RGB-D input. To run it in monocular mode, we use fully-convolutional neural network (FCNN) like \cite{bokovoy2019real} to predict depth of images from camera. Experiments conducted in work \cite{bokovoy2020map} show that RTAB-MAP with CNN-predicted depths is able to successfully build a map of indoor scene in most of cases. Average absolute mapping error was about 0.7m, and most of it was the scale error. After scale correction, average error reduced to 0.3m.

\subsection{Exploration: base version}

\begin{figure}
    \centering
    \includegraphics[width=0.6\textwidth]{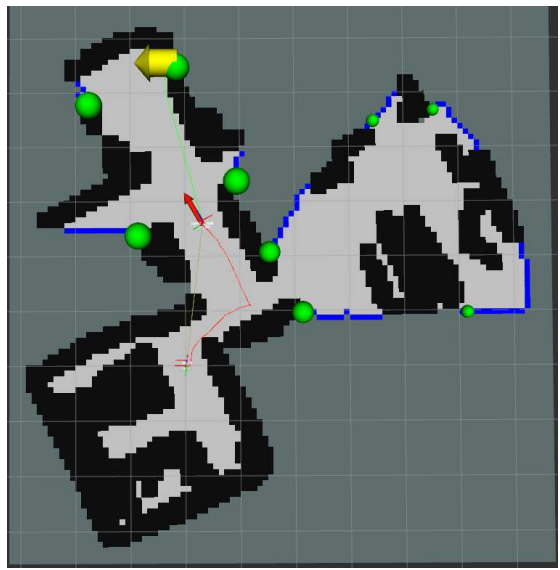}
    \caption{An example of frontier-based exploration work. The black area are occupied map cells, white area are free map cells, blue lines are frontiers, red arrow is robot's pose, and large yellow arrow is the goal chosen by exploration. Green rounds represent cost of frontiers -- the bigger round, the more profitable the frontier}
    \label{figure_frontiers}
\end{figure}

For goal setting, we use frontier-based exploration algorithm \cite{Horner2016} based on \texttt{explore\_lite} ROS package\footnote{http://wiki.ros.org/explore\_lite}. The algorithm from this package looks for frontiers between free and unknown space on 2D SLAM-builded map. To find these frontiers, breadth first search (BFS) in map cell neighborhood graph is used. A centroid of the most "profitable" frontier is marked as goal for robot. An example of frontiers and goal is shown at figure \ref{figure_frontiers}.

Lets describe goal search formally. Let $p \in \mathbb{R}^2$ be current robot position and $F_1, \dots, F_N$ be frontiers found by BFS. Each frontier is represented as set of points on 2D map:

$$F_i = \{f_{i,1}, \dots, f_{i,n_i}\};\ f_{i,j} \in \mathbb{R}^2,\ j=1,\dots,n_i$$

A frontier looks like a chain of cells on occupancy map, i.e. points $f_{i,j}$ and $f_{i,j+1}$ are located in neighbor map cells. Centroid of a frontier is geometrical mean of all its points:

$$centr_i = \frac{1}{n_i} \sum\limits_{j=1}^{n_i} f_{i,j}$$

Frontier cost is a combination of its breadth (i.e. its size in cells) and distance from robot position to it:

\begin{equation}
    cost_i = \alpha ||centr_i - p||_2 - \beta n_i
    \label{eq_frontier_cost_original}
\end{equation}

Breadth is added to cost function with sign "-" because broader frontiers are usually more useful for exploration: the broader frontier is, the more new space we may explore beyond it.

The resultant goal is the centroid of frontier with lowest cost:

\begin{equation}
    goal = centr_{\arg\min\limits_{i} cost_i}
    \label{eq_goal_choice_original}
\end{equation}

The described cost function has significant drawbacks. First, the distance between robot and frontier is measured without obstacle map in mind. So, the path to lowest-cost frontier may be very long that may lead to large exploration time. Second, this cost function does not consider robot's orientation. In context of visual SLAM, large on-the-spot turns may cause localization fails, so the angle that robot shoud turn is also critically important. To eliminate these drawbacks, we modify the cost function and introduced some other enhancements into exploration algorithm. The proposed enhancements are described below.

\subsection{Exploration: our enhancements}

To increase stability and speed of exploration and adapt it to vision-based SLAM methods, we made some enhancements into \texttt{explore\_lite} algorithm. 

First, we changed cost function of frontiers. Instead of euclidean distance between robot and frontier, we used length of robot-frontier path in 2D occupancy map. Also we added orientation - the angle between robot's direction and direction from robot to frontier (i.e. how much should robot turn before it starts moving to the frontier). Our cost estimation formula may be written as follows:

Let $q \in \mathbb{R}^2$ be robot orientation vector, and $\pi = (p_0, p_1, \dots, p_k)$ be path from robot to centroid of $i$-th frontier of size $n_i$. In path $\pi$, $p_0$ is the robot position, and $p_k$ is the centroid of $i-th$ frontier. Our cost function is

\begin{equation}
    \begin{aligned}
    cost_i = \alpha \sum\limits_{j=0}^k ||p_{j+1} - p_j||_2 - \beta n_i + \gamma |\angle(q, p_1 - p_0)|
    \end{aligned}
\end{equation}

where $\alpha$ is coefficient for path length, $\beta$ is coefficient for frontier size, and $\gamma$ is coefficient for turn angle between robot orientation and direction to frontier.

Second, we added some post-processing of SLAM-builded map. We noticed that map constructed by visual SLAM method may contain small fake gaps in obstacles caused by occlusions or low camera resolution. In this case, path planner may find invalid path to goal. So we reduced map resolution from default 0.05m to 0.1m. We performed it by max pooling method with cell type order "unknown $<$ free $<$ occupied". Additionally, we expanded all the obstacles by 1 cell (marked all cells near obstacles as occupied cells).

Also we added a kind of "bump detector" into our exploration algorithm. When path follower sends command "move forward" to robot, and SLAM tracks no forward motion for certain time (e.g. 1 second), we mark map cell ahead robot position as occupied. This trick lets robot not to stuck before invisible obstacle (e.g. small box on the floor in case of tall robot).

Described map post-processing let us significantly decrease amount of fake gaps and non-traversable paths suggested by planner. That makes exploration faster and more stable.

\subsection{Path Planner}

\begin{figure}
    \centering
    \includegraphics[width=0.7\textwidth]{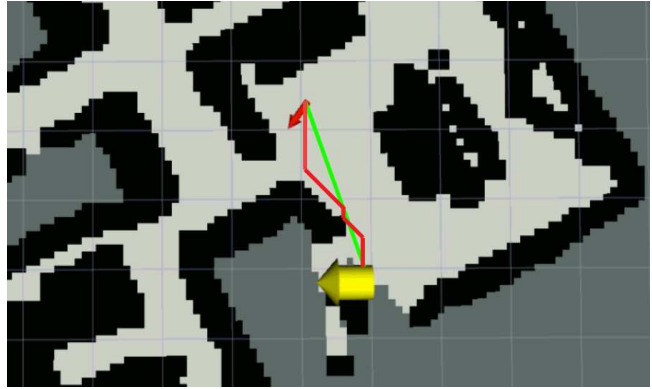}
    \caption{An example of A* path (red) and Theta* path (green)}
    \label{figure_thetastar_path}
\end{figure}

For path planning from robot to goal, we use Theta* algorithm \cite{nash2007theta}. This algorithm has high computational efficiency and supports any-angle paths. So, Theta* paths on occupancy grid are shorter and much smoother than paths of traditional algorithms like Dijkstra \cite{dijkstra1959note} or A* \cite{hart1968formal} (see fig. \ref{figure_thetastar_path}). The path smoothness is critically important for visual SLAM systems because sharp movements may make vSLAM unstable.

The path planner takes post-processed occupancy map, robot and goal positions, and outputs sequence of points that represents robot-goal path. Each point of this sequence is located in free map cell, and each segment between two neighbor points passes through free cells. To track map and pose updates, re-planning is launched with fixed frequency, default 5 Hz. 

\subsection{Path Follower}

To move robot along proposed path, we use simple and straightforward algorithm. We compare robot orientation and direction from robot pose to the next point of the path. If the angle between robot orientation and direction to path point is under some threshold (e.g. less than 5 degrees in absolute value), we move robot forward. If it is above the threshold and is negative, we turn robot left. If it is above the threshold and is positive, we turn robot right.

To increase speed and stability of exploration, we also use some heuristics in path follower. First, at start of exploration, the follower sends only "turn left" command until robot rotates 360 degrees. The "look around" makes exploration faster and sometimes more stable.

Second, in case of vSLAM tracking loss, we launch the following program: rotate 180 degrees left, move a bit forward, and rotate 180 degrees left again. This trick helps robot to return into place stored in SLAM's memory and restore SLAM tracking.

\section{Experiments}
\label{sec:experiments}

\subsection{Experimental setup}

We evaluated our pipeline in both RGB-D and monocular modes. In RGB-D mode, images and precise depths from simulator were sent as input for the SLAM module. In monocular mode, a fully-convolutional neural network (FCNN) was used to estimate depth maps from images. Images with these FCNN-predicted depths were sent as input for SLAM.

\begin{figure}
    \centering
    \includegraphics[width=0.7\textwidth]{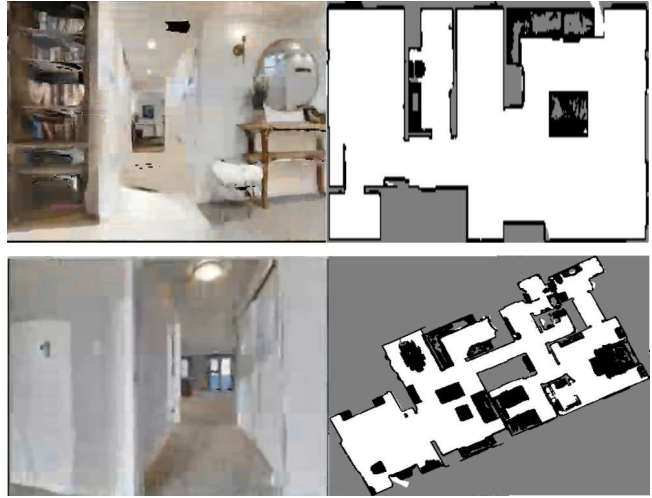}
    \caption{Images and maps of some scenes of Gibson dataset}
    \label{figure_gibson_env}
\end{figure}

We evaluated our exploration pipeline in photo-realistic indoor environment of Habitat simulator \cite{savva2019habitat}. For our experiments we used scenes of Gibson dataset \cite{xia2018gibson}. This dataset was collected in real indoor environments with high-precision Matterport camera\footnote{https://matterport.com/} and accurate algorithmic post-processing. That let us receive realistic image and precise depth map from each point of scene.

Gibson dataset contains about 500 scenes. Most of them are apartments or living houses. Many scenes have defects like gaps in textures that may cause exploration fails and incorrect quality estimation. Also, many scenes have several floors, so 2D SLAM does not work on them. Therefore we selected only 31 scenes for our experiments -- the scenes without stairs and texture defects. Area of selected scenes varied from 28 to 251 $m^2$. An example of images and maps of these scenes is shown in Fig. \ref{figure_gibson_env}.

To measure exploration efficiency, we used coverage metric - the area of map space explored at certain time. We measured both absolute and relative area. The area values were measured for different time from start -- from 15s to 240s. 

Another efficiency metric that we computed was the number of scenes where exploration was finished in 240 seconds, and average finish time on these scenes. We considered exploration as finished when explored area was more than 95\% of total scene area.

Also we counted number of SLAM tracking losses over all the scenes to measure exploration stability.

\subsection{Results with RGB-D input}

For broad evaluation of the whole exploration pipeline and our enhancements in RGB-D mode, we carried out a set of experiments on selected scenes with precise depths at SLAM module input. First, we tested our pipeline with unchanged \texttt{explore\_lite} algorithm as exploration module. Second, we added a "bump detector" and tested our pipeline again. Next, we added obstacle expanding into map post-processing to test its effect. And finally, we included the last our enhancement - added orientation coefficient into cost function. To make metric values more stable, we launched exploration with each enhancement 5 times on all scenes, and averaged metric values through these 5 tests.

To examine behaviour of our exploration in both large and small environments, we selected 13 relatively large scenes (with area more than 60 $m^2$), and 18 scenes with area under 60 $m^2$. We estimated coverage metrics on large and small scenes separately.

\begin{figure}
    \centering
    \includegraphics[width=0.9\textwidth]{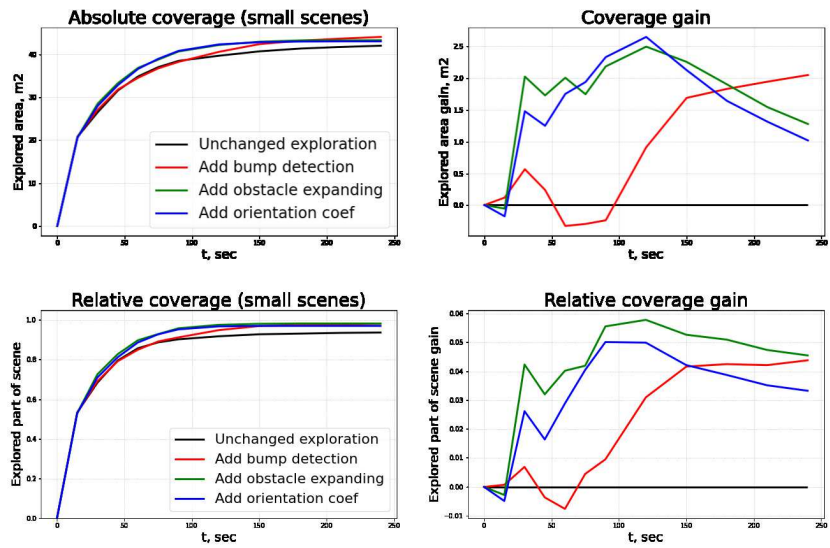}
    \caption{Coverage of exploration on small scenes: top left -- absolute coverage values, top right - absolute coverage gain (compared to unchanged \texttt{explore\_lite} algorithm, bottom left -- coverage relative to total scene area, bottom right -- gain of relative coverage (compared to unchanged \texttt{explore\_lite} algorithm)}
    \label{figure_metrics_small}
\end{figure}

\begin{figure}
    \centering
    \includegraphics[width=0.9\textwidth]{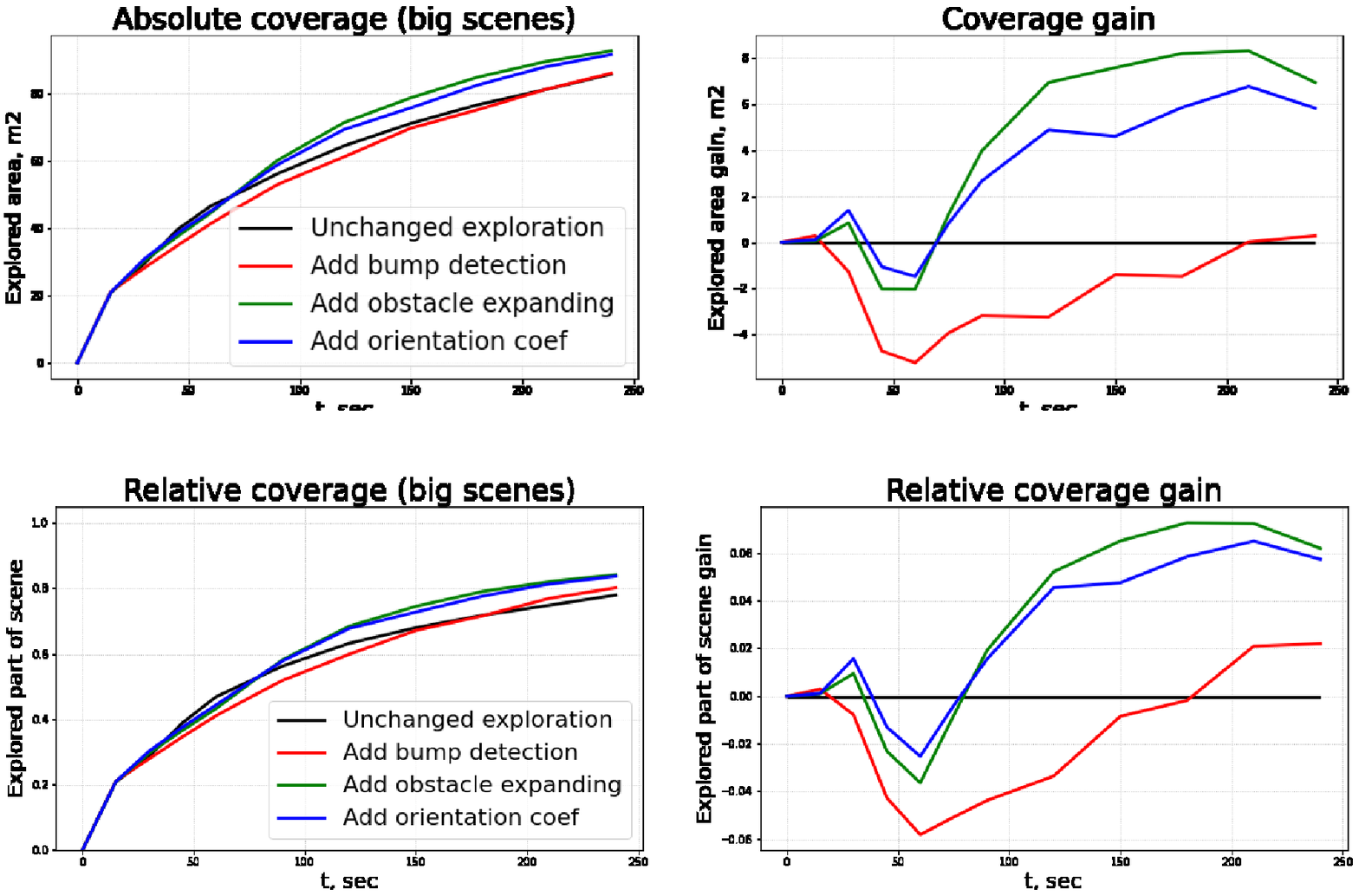}
    \caption{Coverage of exploration on large scenes: top left -- absolute coverage values, top right -- absolute coverage gain (compared to unchanged \texttt{explore\_lite} algorithm, bottom left - coverage relative to total scene area, bottom right -- gain of relative coverage (compared to unchanged \texttt{explore\_lite} algorithm}
    \label{figure_metrics_large}
\end{figure}

\begin{figure}
    \centering
    \includegraphics[width=0.9\textwidth]{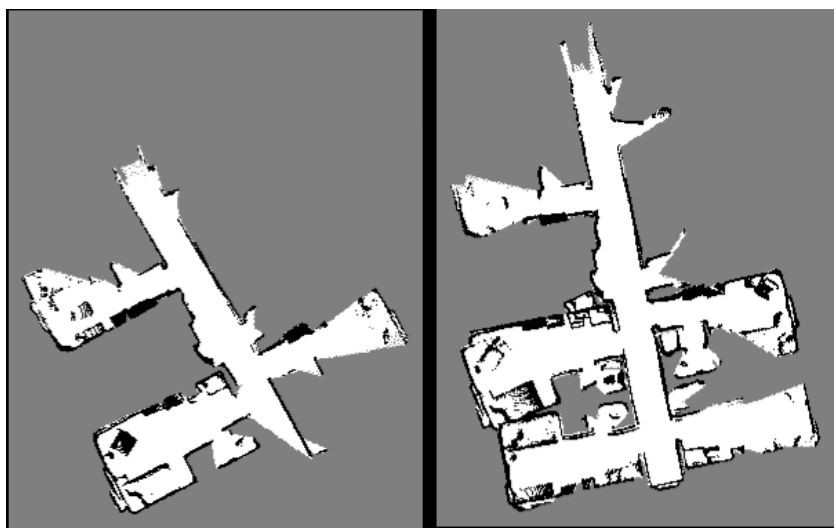}
    \caption{Maps builded by SLAM during 240 seconds of exploration on a large scene. Left -- with unchanged exploration, right -- with all our enhancements}
    \label{figure_slam_map_comparison}
\end{figure}

The coverage results on small scenes are shown at fig. \ref{figure_metrics_small}. At full experiment time (240 seconds), unchanged \texttt{explore\_lite} algorithm covered about 94\% of scene area at average. With adding bump detection, 240-second covered area increased to about 97\%. Obstacle expanding and orientation coefficient had no significant effect to total covered area, but had significant positive effect to exploration speed - the area covered in 90 seconds increased from 91\% to 96\%.

The coverage values for exploration on large scenes are shown at fig. \ref{figure_metrics_large}. Differences at large scenes were more noticeable than on small ones (see for example fig. \ref{figure_slam_map_comparison}). With unchanged exploration, total covered area was at average 78\% in relative value and 85 $m^2$ in absolute value. With adding an imitation of "bump detector", the covered area increased to 80\% and 86 $m^2$ respectfully. With adding obstacle expanding, the coverage increased to 84\% and 92.5 $m^2$. But adding orientation coefficient made no progress in coverage - explored area remained at level of 84\%.

\begin{table}[]
    \centering
    \begin{tabular}{|c|c|c|c|}
        \hline
        Enhancement  & N of SLAM losses & N of finished scenes & Avg. finish time, s\\
        \hline
        No & 14.2 & 21.6 & 213\\
        Bump detection & 13.8 & 21.75 & 205\\
        Obstacle expanding & 10.0 & 23.4 & 165\\
        Orientation coef & 10.4 & 23.2 & 157\\
        \hline
        
    \end{tabular}
    \caption{Stability and efficiency metric values for exploration with different enhancements}
    \label{table_evaluation}
\end{table}

The results of SLAM stability and exploration efficiency evaluation are shown in table \ref{table_evaluation}. Bump detection and obstacle expanding had great positive effect to all of the metrics. Adding orientation coefficient to cost function did not influence SLAM stability, but reduced a bit average finish time - from 165s to 157s. Such weak effect of orientation coefficient may be probably caused by large amount of "dead ends" in scenes selected for evaluation. So, robot had to turn around many times in these dead ends regardless of frontier cost function. This hypothesis may be checked by experiments on large scenes with spacious rooms.

Overall, proposed enhancements of exploration algorithm improved quality for all of estimated metrics. With introduction of a kind of "bump detector", obstacle expanding, and orientation coefficient of cost-function, average part of explored area increased by 3\% on small scenes and 6\% on large scenes. Also these enhancements reduced number of SLAM losses by almost 1.5 times and reduced average exploration finish time from 213 seconds to 157 seconds.

\subsection{Results with FCNN-predicted depths}

\begin{figure}
    \centering
    \includegraphics[width=0.8\textwidth]{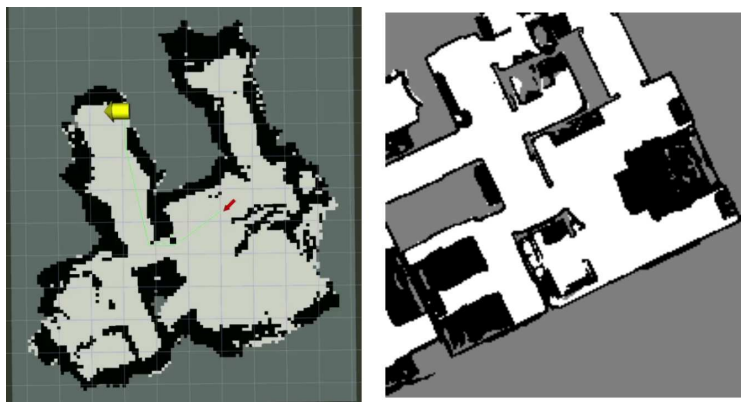}
    \caption{An example of map constructed during exploration with FCNN depths (left) compared to ground-truth map (right)}
    \label{figure_fcnn_map_example}
\end{figure}

\begin{figure}
    \centering
    \includegraphics[width=0.8\textwidth]{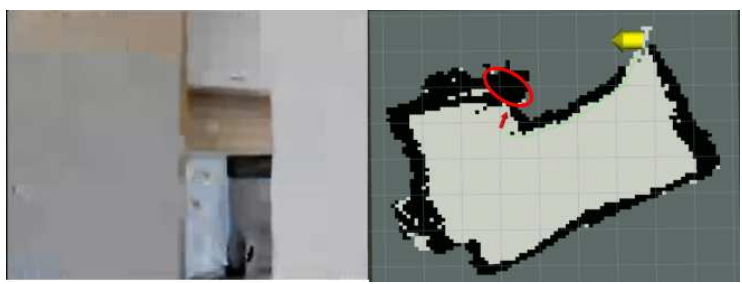}
    \caption{Example of inaccurate mapping with FCNN-predicted depths. On the camera view (left part of picture), small doorway is observed. But on the map (right part of the picture), the doorway is marked as a wall (see red ellipsis)}
    \label{figure_doorway_ignore}
\end{figure}

\begin{figure}
    \centering
    \includegraphics[width=0.9\textwidth]{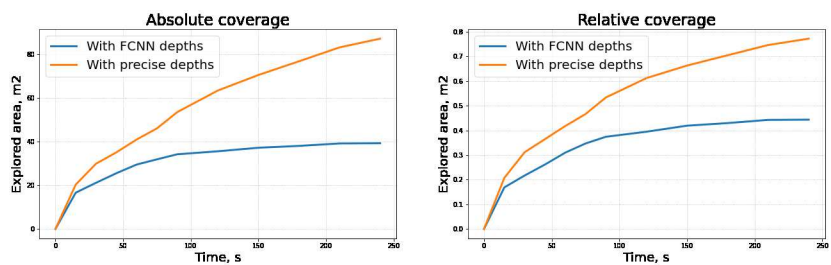}
    \caption{Comparison of exploration with FCNN-predicted depths and exploration with precise depths: absolute coverage (left), and relative scene coverage (right)}
    \label{figure_fcnn_metrics}
\end{figure}

To examine our pipeline in monocular mode, we carried out some experiments on selected scenes. We used pre-trained network from \cite{bokovoy2019real} for depth prediction. To adapt this network to simulated environment and Habitat camera parameters, we fine-tuned it on approximately 38000 image-depth pairs from 25 scenes of our selected collection. The other 6 selected scenes were used for evaluation of the exploration pipeline.

The experiments showed that our exploration pipeline is able to work autonomously and build plausible map (see Fig. \ref{figure_fcnn_map_example}) in monocular mode. However, errors in neural depth estimation caused some errors in constructed map. For example, a narrow doorways sometimes were mapped as continuous wall (see Fig. \ref{figure_doorway_ignore}). Due to such inaccurate mapping, the planner could not find paths to far goals, and exploration algorithm explored only part of scene area. Coverage metric values are shown in Fig. \ref{figure_fcnn_metrics}. Average explored part of area reached 44\% (compared to 77\% with exploration in RGB-D mode).

Overall, our tests showed that proposed exploration pipeline is able to work in monocular mode with neural depth estimation, but inaccurate depth prediction may lead to mapping errors and incomplete area coverage. These errors may be eliminated with more thorough neural network fine-tuning and fine adjustment of SLAM parameters. A video with demonstration of exploration with our enhancements and FCNN-predicted depths is available at \\
 \url{https://drive.google.com/file/d/1QJWmjR9Y2VWbycZVwz3Y6Dl9Rzkp-zjB/view?usp=sharing}.

\section{Conclusion and future work}
\label{sec:conclusion}

We introduced novel enhancements to exploration algorithm and evaluated them in photo-realistic simulated environment. We showed that our enhancements increase the area of the explored space, reduce the time needed for full scene exploration and reduce number of tracking losses with vSLAM operating ground-truth depth map. We also tested our approach in monocular mode, with FCNN-predicted depth maps. The results show that the exploration algorithm with our enhancements is able to explore about a half of environment in monocular mode.

In future we plan to carry out more research into monocular vSLAM to increase its accuracy and exploration coverage. Possible ways of increasing vSLAM quality are usage of novel time-consistent FCNN architectures, global depth correction with geometric methods, and thorough vSLAM map post-processing.


%
%
\bibliographystyle{splncs04}
\bibliography{main}
%




\end{document}